# Cross-lingual Annotation Projection of Semantic Roles

**Sebastian Padó**　　　　　　　　　　　　　　　　PADO@IMS.UNI-STUTTGART.DE
*Institut für maschinelle Sprachverarbeitung*
*Universität Stuttgart, 70174 Stuttgart, Germany*

**Mirella Lapata**　　　　　　　　　　　　　　　　MLAP@INF.ED.AC.UK
*School of Informatics, University of Edinburgh*
*10 Crichton Street, Edinburgh EH8 10 AB, UK*

## Abstract

This article considers the task of automatically inducing role-semantic annotations in the FrameNet paradigm for new languages. We propose a general framework that is based on annotation projection, phrased as a graph optimization problem. It is relatively inexpensive and has the potential to reduce the human effort involved in creating role-semantic resources. Within this framework, we present projection models that exploit lexical and syntactic information. We provide an experimental evaluation on an English-German parallel corpus which demonstrates the feasibility of inducing high-precision German semantic role annotation both for manually and automatically annotated English data.

## 1. Introduction

Semantic roles play a prominent role in linguistic theory (Fillmore, 1968; Jackendoff, 1990; Dowty, 1991). They describe the relations that hold between a predicate and its arguments, abstracting over surface syntactic configurations. As an example, consider sentences (1a) and (1b) where *butter* is uniformly assigned the semantic role UNDERGOER (since it undergoes a physical change) even though it is syntactically realized as the object of the verb *melt* in (1a) and its subject in (1b):

(1)　　　a.　[Bob]ₐGENT **melted** [the butter]UNDERGOER.
　　　　　b.　[The butter]UNDERGOER **melted**.

This intermediate representation seems a promising first step towards text understanding, and can ultimately benefit many natural language processing tasks that require broad coverage semantic processing.

Methods for the automatic identification and labeling of semantic roles, often referred to as *shallow semantic parsing* (Gildea & Jurafsky, 2002), are an important prerequisite for the widespread use of semantic role information in large-scale applications. The development of shallow semantic parsers[1] has been greatly facilitated by the availability of resources like FrameNet (Fillmore, Johnson, & Petruck, 2003) and PropBank (Palmer, Gildea, & Kingsbury, 2005), which document possible surface realization of semantic roles. Indeed, semantic roles have recently found use in applications ranging from information extraction (Surdeanu, Harabagiu, Williams, & Aarseth, 2003) to the modeling of textual entailment relations (Tatu & Moldovan, 2005; Burchardt & Frank, 2006), text categorization (Moschitti, 2008), question answering (Narayanan & Harabagiu, 2004; Frank, Krieger, Xu, Uszkoreit, Crysmann, Jörg, & Schäfer, 2007; Moschitti, Quarteroni, Basili, & Manandhar, 2007; Shen & Lapata, 2007), machine translation (Wu & Fung, 2009a, 2009b) and its evaluation (Giménez & Màrquez, 2007).

---

1. Approaches to building shallow semantic parsers are too numerous to list. We refer the interested reader to the proceedings of the 2005 CoNLL shared task (Carreras & Màrquez, 2005) and to the 2008 Computational Linguistics Special Issue on Semantic Role Labeling (Màrquez, Carreras, Litkowski, & Stevenson, 2008) for an overview of the state-of-the-art.

| DEPARTING |
|---|
| An object (the THEME) moves away from a SOURCE. |

|  |  |
|---|---|
| THEME | **The officer** left the house. |
|  | **The plane** leaves at seven. |
|  | **His** departure was delayed. |
| SOURCE | We departed **from New York**. |
|  | He retreated **from his opponent**. |
|  | The woman left **the house**. |
| FEEs | abandon.v, desert.v, depart.v, departure.n, emerge.v, emigrate.v, emigration.n, escape.v, escape.n, leave.v, quit.v, retreat.v, retreat.n, split.v, withdraw.v, withdrawal.n |

Table 1: Abbreviated FrameNet entry for the DEPARTING frame

In the FrameNet paradigm, the meaning of a predicate (usually a verb, noun, or adjective) is represented by reference to a *frame*, a prototypical representation of the situation the predicate describes (Fillmore, 1982). The semantic roles, which are called *frame elements*, correspond to entities present in this situation, and are therefore frame-specific. For each frame, the English FrameNet database[2] lists the predicates that can evoke it (called *frame-evoking elements* or FEEs), gives the possible syntactic realizations of its semantic roles, and provides annotated examples from the British National Corpus (Burnard, 2000). An abbreviated example definition of the DEPARTING frame is shown in Table 1. The semantic roles are illustrated with example sentences and the FEEs are shown at the bottom of the table (e.g., *abandon, desert, depart*). The PropBank corpus, the second major semantic role resource for English, provides role realization information for verbs in a similar manner on the Wall Street Journal portion of the Penn Treebank. It uses index-based role names (Arg0–Arg$n$), where Arg0 and Arg1 correspond to Dowty's (1991) proto-AGENT and proto-PATIENT. Higher indices are defined on a verb-by-verb basis.

Unfortunately, resources such as FrameNet and PropBank are largely absent for almost all languages except English, the main reason being that role-semantic annotation is an expensive and time-consuming process. The current English FrameNet (Version 1.3) has been developed over the past twelve years. It now contains roughly 800 frames covering around 150,000 annotated tokens of 7,000 frame-evoking elements. Although FrameNets are being constructed for German, Spanish, and Japanese, these resources are considerably smaller. The same is true for PropBank-style resources, which have been developed for Korean[3], Chinese (Xue & Palmer, 2009), Spanish and Catalan (Taulé, Mart, & Recasens, 2008). Compared to the English PropBank, which covers 113,000 predicate-argument structures, the resources for the other languages are two to three times smaller (e.g., the Korean PropBank provides 33,000 annotations).

---

2. Available from `http://framenet.icsi.berkeley.edu`.

Given the data requirements for supervised learning algorithms (Fleischman & Hovy, 2003) and the current paucity of such data, unsupervised methods could potentially enable the creation of annotated data for new languages and reduce the human effort involved. However, unsupervised approaches to shallow semantic parsing are still at an early stage, and mostly applicable to resources other than FrameNet (Swier & Stevenson, 2004, 2005; Grenager & Manning, 2006). In this article, we propose a method which employs parallel corpora for acquiring frame elements and their syntactic realizations for new languages (see the upper half of Table 1). Our approach leverages the existing English FrameNet to overcome the resource shortage in other languages by exploiting the translational equivalences present in aligned data. Specifically, it uses *annotation projection* (Yarowsky, Ngai, & Wicentowski, 2001; Yarowsky & Ngai, 2001; Hwa, Resnik, Weinberg, & Kolak, 2002; Hi & Hwa, 2005) to transfer semantic roles from English to less resource-rich languages. The key idea of projection can be summarized as follows: (1) given a pair of sentences $E$ (English) and $L$ (new language) that are translations of each other, annotate $E$ with semantic roles; and then (2) project these roles onto $L$ using word alignment information. In this manner, we induce semantic structure on the $L$ side of the parallel text, which can then serve as data for training a shallow semantic parser for $L$ that is independent of the parallel corpus.

The annotation projection paradigm faces at least two challenges when considering semantic roles. Firstly, the semantic structure to be projected must be shared between the two sentences. Clearly, if the role-semantic analysis of the source sentence $E$ is inappropriate for the target sentence $L$, simple projection will not produce valid semantic role annotations. Secondly, even if two sentences demonstrate semantic parallelism, semantic role annotations pertain to potentially arbitrarily long word spans rather than to individual words. Recovering the word span of the semantic roles in the target language is challenging given that automatic alignment methods often produce noisy or incomplete alignments.

We address the first challenge by showing that, if two languages exhibit a substantial degree of semantic correspondence, then annotation projection is feasible. Using an English-German parallel corpus as a test bed, we assess whether English semantic role annotations can be transferred successfully onto German. We find that the two languages exhibit a degree of semantic correspondence substantial enough to warrant projection. We tackle the second challenge by presenting a framework for the projection of semantic role annotations that goes beyond single word alignments. Specifically, we construct *semantic alignments* between constituents of source and target sentences and formalize the search for the best semantic alignment as an optimization problem in a bipartite graph. We argue that bipartite graphs offer a flexible and intuitive framework for modeling semantic alignments that is able to deal with noise and to represent translational divergences. We present different classes of models with varying assumptions regarding admissible correspondences between source and target constituents. Experimental results demonstrate that constituent-based models outperform their word-based alternatives by a large margin.[4]

---

3. The Korean PropBank is available from the LDC (`http://www.ldc.upenn.edu/`).

The remainder of this article is organized as follows. Section 2 discusses annotation projection in general and presents an annotation study examining the degree of semantic parallelism on an English-German corpus. In Section 3, we formalize semantic alignments and present our modeling framework. Our experiments are detailed in Section 4. We review related work in Section 5 and conclude the article with discussion of future work (Section 6).

## 2. Annotation Projection and Semantic Correspondence

In recent years, interest has grown in parallel corpora for multilingual and cross-lingual natural language processing. Beyond machine translation, parallel corpora can be exploited to relieve the effort involved in creating annotations for new languages. One important paradigm, *annotation projection*, creates new *monolingual* resources by transferring annotations from English (or other resource-rich languages) onto resource-scarce languages through the use of word alignments. The resulting (noisy) annotations can be then used in conjunction with robust learning algorithms to obtain NLP tools such as taggers and chunkers relatively cheaply. The projection approach has been successfully used to transfer a wide range of linguistic annotations between languages. Examples include parts of speech (Yarowsky et al., 2001; Hi & Hwa, 2005), chunks (Yarowsky et al., 2001), dependencies (Hwa et al., 2002), word senses (Diab & Resnik, 2002; Bentivogli & Pianta, 2005), information extraction markup (Riloff, Schafer, & Yarowsky, 2002), coreference chains (Postolache, Cristea, & Orasan, 2006), temporal information (Spreyer & Frank, 2008), and LFG f-structures (Tokarczyk & Frank, 2009).

An important assumption underlying annotation projection is that linguistic analyses in one language will be also valid in another language. It is however unrealistic to expect any two languages, even of the same family, to be in perfect correspondence. There are many well-studied systematic differences across languages often referred to as *translational divergences* (van Leuven-Zwart, 1989; Dorr, 1995). These can be structural, where the same semantic content in the source and in the target language can be realized using different structures, or semantic, where the content itself undergoes a change in translation. Translational divergences (in conjunction with poor alignments) are a major stumbling block towards achieving accurate projections. Yarowsky and Ngai (2001) find that parts of speech that are transferred directly from English onto French contain considerable noise, even in cases where inaccurate automatic alignments have been manually corrected (accuracies vary between 69% and 78% depending on tagset granularity). For syntax, Hwa, Resnik, Weinberg, Cabezas, and Kolak (2005) find that only 37% of English dependency relations have direct counterparts in Chinese, and 38% in Spanish. The problem is commonly addressed with filtering mechanisms, which act as a post-processing step on the projection output. For example, Yarowsky and Ngai (2001) exclude infrequent projections and poor alignments and Hwa et al. (2005) apply transformation rules which encode linguistic knowledge about the target language.

---

4. A preliminary version of this work was published in the proceedings of EMNLP 2005 and COLING/ACL 2006. The current article contains a more detailed description of our approach, presents several novel experiments, and a comprehensive error analysis.

In the case of semantic frames there is reason for optimism. By definition, frames are based on conceptual structure (Fillmore, 1982). The latter constitute generalizations over surface structure and therefore ought to be less prone to syntactic variation. Indeed, efforts to develop FrameNets manually in German, Japanese, and Spanish reveal that a large number of English frames can be re-used directly to describe predicates and their arguments in other languages (Ohara, Fujii, Saito, Ishizaki, Ohori, & Suzuki, 2003; Subirats & Petruck, 2003; Burchardt, Erk, Frank, Kowalski, Padó, & Pinkal, 2009). Boas (2005) even suggests frame semantics as an interlingual meaning representation.

Computational studies on projection in parallel corpora have also obtained good results for semantic annotation. Fung and Chen (2004) induce FrameNet-style annotations in Chinese by mapping English FrameNet entries directly onto concepts listed in HowNet[5], an on-line ontology for Chinese, without using parallel texts. In their experiment, they transfer semantic roles from English to Chinese with an accuracy of 68%. Basili, Cao, Croce, Coppola, and Moschitti (2009) use gold standard annotations to transfer semantic roles from English to Italian with 73% accuracy. Bentivogli and Pianta (2005) project EuroWordNet sense tags, which represent more fine-grained semantic information than FrameNet, also from English to Italian. They obtain a precision of 88% and a recall of 71%, without applying any filtering. Fung, Wu, Yang, and Wu (2006, 2007) analyse an automatically annotated English–Chinese parallel corpus and find high cross-lingual agreement for PropBank roles (in the range of 75%–95%, depending on the role).

To provide a sound empirical justification for our projection-based approach, we conducted a manual annotation study on a parallel English-German corpus. We identified semantic role information in bi-sentences and assessed the degree to which frames and semantic roles agree and diverge in English and German. The degree of divergence provides a natural upper bound for the accuracy attainable with annotation projection.

## 2.1 Sample Selection

English-German bi-sentences were drawn from the second release of Europarl (Koehn, 2005), a corpus of professionally translated proceedings of the European Parliament. Europarl is aligned at the document and sentence level and is available in 11 languages. The English–German section contains about 25 million words on both sides. Even though restricted in genre (transcriptions of spoken text), Europarl is fairly open-domain, covering a wide range of topics such as foreign politics, cultural and economic affairs, and procedural matters.

A naive sampling strategy would involve randomly selecting bi-sentences from Europarl which contain a FrameNet predicate on the English side aligned to some word on the German side. There are two caveats here. First, the alignment between the two predicates may be wrong, leading us to assign a wrong frame to the German predicate. Secondly, even if the alignment is accurate, it is possible that a randomly chosen English predicate evokes a frame that is not yet covered by FrameNet. For example, FrameNet 1.3 documents the "receive" sense of the verb *accept* (as in the sentence *Mary accepted a gift*), but has no entry for the "admit" sense of the predicate (e.g., in *I accept that this is a problem for*

---

5. See `http://www.keenage.com/zhiwang/e_zhiwang.html`.

| Measure | English | German | All |
|---|---|---|---|
| Frame Match | 89.7 | 86.7 | 88.2 |
| Role Match | 94.9 | 95.2 | 95.0 |
| Span Match | 84.4 | 83.0 | 83.7 |

Table 2: Monolingual inter-annotator agreement on the calibration set

*the EU*) which is relatively frequent in Europarl. Indeed, in a pilot study, we inspected a small random sample consisting of 100 bi-sentences, using the publicly available GIZA++ software (Och & Ney, 2003) to induce English-German word alignments. We found that 25% of the English predicates did not have readings documented in FrameNet, and an additional 9% of the predicate pairs were instances of wrong alignments. In order to obtain a cleaner sample, our final sampling procedure was informed by the English FrameNet and SALSA, a FrameNet-compatible database, for German (Erk, Kowalski, Padó, & Pinkal, 2003).

We gathered all German–English sentences in the corpus that had at least one pair of GIZA++-aligned predicates $(w_e, w_g)$, where $w_e$ was listed in FrameNet and $w_g$ in SALSA, and where the intersection of the two frame lists for $w_g$ and $w_e$ was non-empty. This corpus contains 83 frame types, 696 lemma pairs, and 265 unique English and 178 unique German lemmas. Sentence pairs were grouped into three bands according to their frame frequency (High, Medium, Low). We randomly selected 380 pairs from each band for annotation. The total sample consisted of 1,140 bi-sentences. Before semantic annotation took place, constituency parses for the corpus were obtained from Collins' (1997) parser for English and Dubey's (2005) for German. The automatic parses were then corrected manually, following the annotation guidelines of the Penn Treebank (English) and the TIGER corpus (German).

## 2.2 Annotation

After syntactic correction, two annotators with native-level proficiency in German and English annotated each bi-sentence with the frames evoked by $w_e$ and $w_g$ and their semantic roles (i.e., one frame per monolingual sentence). For every predicate, the task involved two steps: (a) selecting the appropriate frame and (b) assigning the instantiated semantic roles to sentence constituents. Annotators were provided with detailed guidelines that explained the task with multiple examples.

The annotation took place on the gold standard parsed corpus and proceeded in three phases: a training phase (40 bi-sentences), a calibration phase (100 bi-sentences), and a production mode phase (1000 bi-sentences). During training, annotators were acquainted with the annotation style. In the calibration phase, each bi-sentence was doubly annotated to assess inter-annotator agreement. Finally, in production mode, each of the 1000 bi-sentences in the main dataset was split and each half randomly assigned to one of the coders for single annotation. We thus ensured that no annotator saw both parts of any bi-sentence to avoid any language bias in the role assignment (annotators may be prone to label an English sentence similar to its German translation and vice versa). Each coder annotated approximately the same amount of data in English and German and had access to the FrameNet and SALSA resources.

| Measure | Precision | Recall | F1-Score |
|---|---|---|---|
| Frame Match | 71.6 | 71.6 | 71.6 |
| Role Match | 90.5 | 92.3 | 91.4 |

Table 3: Semantic parallelism between English and German

The results of our inter-annotator agreement study are given in Table 2. The widely used Kappa statistic is not directly applicable to our task as it requires a fixed set of items to be classified into a fixed set of categories. In our case, however, there are no fixed items, since the span for the frame elements can have any length. In addition, the categories (i.e., frames and roles) are predicate-specific, and vary from item to item (for a discussion of this issue, see also the work of Miltsakaki et al., 2004). Instead, we compute three different agreement measures defined as: the ratio of common frames between two sentences (Frame Match), the ratio of common roles (Role Match), and the ratio of roles with identical spans (Span Match). As shown in Table 2, annotators tend to agree in frame assignment; disagreements are mainly due to fuzzy distinctions between closely related frames (e.g., between AWARENESS and CERTAINTY). Annotators also agree on what roles to assign and on identifying role spans. Overall, we obtain high agreement for all aspects of the annotation, which indicates that the task is well-defined. We are not aware of published agreement figures on the English FrameNet annotations, but our results are comparable to numbers reported by Burchardt, Erk, Frank, Kowalski, Padó, and Pinkal (2006) for German, viz. 85% agreement on frame assignment (Frame Match) and 86% agreement on role annotation.[6]

## 2.3 Evaluation

Recall that the main dataset consists of 1,000 English-German bi-sentences annotated with FrameNet semantic roles. Since the annotations for each language have been created independently, they can be used to provide an estimate of the degree of semantic parallelism between the two languages. We measured parallelism using precision and recall, treating the German annotations as gold standard. This evaluation scheme directly gauges the usability of English as a source language for annotation projection. Less than 100% recall means that the target language has frames or roles which are not present in English and cannot be retrieved by annotation projection. Conversely, imperfect precision indicates that there are English frames or roles whose projection yields erroneous annotations in the target language. Frames and roles are counted as matching if they occur in both halves of a bi-sentence, regardless of their role spans, which are not comparable across languages.

As shown in Table 3, about 72% of the time English and German evoke the same frame (Frame Match). This result is encouraging, especially when considering that frame disagreements also arise within a single language as demonstrated by our inter-annotator study on the calibration set (see the row Frame Match in Table 2). However, it also indicates that there is a non-negligible number of cases of translational divergence on the frame level. These are often cases where one language chooses a single predicate to express a situation whereas the other one uses complex predication. In the following example, the English transitive predicate *increase* evokes the frame CAUSE_CHANGE_OF_SCALAR_POSITION ("An agent or cause increases the position of a variable on some scale"). Its German translation *führt zu höheren* ("leads to higher") combines the CAUSATION frame evoked by *führen* with the inchoative CHANGE_OF_SCALAR_POSITION frame introduced by *höher*:

(2)     This will **increase** the level of employment.

---

6. The parallel corpus we created is available from `http://nlpado.de/~sebastian/srl_data.html`.

        Dies wird zu einer **höheren** Erwebsquote          **führen**.
        This will  to a      higher    level of employment lead

At the level of semantic roles, agreement (Role Match) reaches an F1-Score of 91%. This means that when frames correspond across languages, the roles agree to a large extent. Role mismatches are frequently cases of passivization or infinitival constructions leading to role elision. In the example below, *remembered* and *denkt* both evoke the MEMORY frame. English uses a passive construction which leaves the deep subject position unfilled. In contrast, German uses an active construction where the deep subject position is filled by a semantically light pronoun, *man* ("one").

(3)     I ask that [Ireland]$_{\text{CONTENT}}$ be **remembered**.

        Ich möchte    darum bitten, dass [man]$_{\text{COGNIZER}}$ [an Irland]$_{\text{CONTENT}}$ **denkt**.
        I    would like      to ask that one          of Ireland        thinks

In sum, we find that there is substantial cross-lingual semantic correspondence between English and German provided that the predicates evoke the same frame. We enlisted the help of the SALSA database to meet this requirement. Alternatively, we could have used an existing bilingual dictionary (Fung & Chen, 2004), aligned the frames automatically using a vector-based representation (Basili et al., 2009) or inferred FrameNet-style predicate labels for German following the approach proposed by Padó and Lapata (2005).

## 3. Modeling Semantic Role Projection with Semantic Alignments

Most previous work on projection relies on word alignments to transfer annotations between languages. This is not surprising, since the annotations of interest are often defined at the word level (e.g., parts of speech, word senses, or dependencies) and rarely span more than one token. In contrast, semantic roles can cover sentential constituents of arbitrary length, and simply using word alignments for projection is likely to result in wrong role spans.

As an example, consider the bi-sentence in Figure 1.[7] Assume for now that the (English) source has been annotated with semantic roles which we wish to project onto the (German) target. Although some alignments (indicated by the dotted lines below the sentence) are accurate (e.g., *promised* ↔ *versprach*, *to* ↔ *zu*), others are noisy or incomplete (e.g., *time* ↔ *pünktlich* instead of *on time* ↔ *pünktlich*). Based on these alignments, the MESSAGE role would be projected into German onto the (incorrect) word span *pünktlich zu* instead of *pünktlich zu kommen*, since *kommen* is not aligned with any English word.

It is of course possible to devise heuristics for amending alignment errors. However, this process does not scale well: different heuristics need to be created for different errors, and the process has to be repeated for each new language pair. Instead, our projection model alleviates this problem in a more principled manner by taking constituency information into account. Specifically, we induce *semantic alignments* between source and target sentences by relying on syntactic constituents to introduce a bias towards linguistically meaningful spans. For a constituent to be aligned correctly, it is sufficient that a subset of its yield is correctly word-aligned. So, in Figure 1, we can align *to be on time* with *pünktlich zu kommen* and project the role MESSAGE accurately, despite the fact that *be* and *kommen* are not aligned with each other. In the following, we describe in detail how semantic alignments are computed and subsequently guide projection onto the target language.

---

7. A literal gloss of the German sentence is *Kim promises timely to come*.

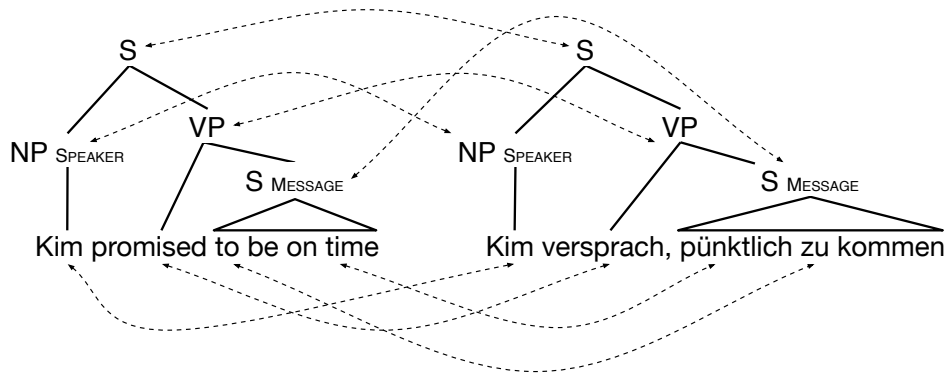

Figure 1: Bilingual projection of semantic role information with semantic alignments between constituents.

### 3.1 Framework Formalization

Each bi-sentence is represented as a set of linguistic units. These are distinguished into source ($u_s \in U_s$) and target ($u_t \in U_t$) units and can be words, chunks, constituents, or other groupings. The semantic roles for the source sentence are modeled as a labeling function $a_s : R \to 2^{U_s}$ which maps roles to sets of source units. We view projection as the construction of a similar role labeling function on the target sentence, $a_t : R \to 2^{U_t}$. Without loss of generality, we limit ourselves to one frame per sentence, as does FrameNet.[8]

A *semantic alignment $A$* between $U_s$ and $U_t$ is a subset of the Cartesian product of the source and target units:

$$A \subseteq U_s \times U_t \tag{4}$$

An alignment link between $u_s \in U_s$ and $u_t \in U_t$ implies that $u_s$ and $u_t$ are semantically equivalent. Provided with $A$ and the role assignment function for the source sentence, $a_s$, projection consists simply of transferring the source labels $r$ onto the union of the target units that are semantically aligned with the source units bearing the label $r$:

$$a_t(r) = \{u_t \, \| \, \exists \, u_s \in a_s(r) : (u_s, u_t) \in A\} \tag{5}$$

---

8. This entails that we cannot take advantage of potentially beneficial dependencies between the arguments of different predicates within one sentence, which have been shown to improve semantic role labeling (Carreras, Màrquez, & Chrupała, 2004).

We phrase the search for a semantic alignment $A$ as an *optimization problem*. Specifically, we seek the alignment that maximizes the product of bilingual similarities $sim$ between source and target units:

$$\hat{A} = \operatorname*{argmax}_{A \in \mathcal{A}} \prod_{(u_s, u_t) \in A} sim(u_s, u_t) \tag{6}$$

There are several well-established methods in the literature for computing semantic similarity within one language (see the work of Weeds, 2003, and Budanitsky & Hirst, 2006, for overviews). Measuring semantic similarity across languages is not as well studied and there is less consensus on what the appropriate methods are. In this article, we employ a very simple method, using automatic word alignments as a proxy for semantic equivalence; however, other similarity measures can be used (see the discussion in Section 6). Following general convention, we assume that $sim$ is a function ranging from 0 (minimal similarity) to 1 (maximal similarity).

A wealth of optimization methods can be used to solve (6). In this article, we treat constituent alignment as a *bipartite graph* optimization problem. Bipartite graphs provide a simple and intuitive modeling framework for alignment problems and their optimization algorithms are well-understood and computationally moderate. More importantly, by imposing constraints on the bipartite graph, we can bias our model against linguistically implausible alignments, for example alignments that map multiple English roles onto a single German constituent. Different graph topologies correspond to different constraints on the set of *admissible alignments* $\mathcal{A}$. For instance, we may want to ensure that all source and target units are aligned, or to restrict alignment to one-to-one matches (see Section 3.3 for further details).

More formally, a weighted bipartite graph is a graph $G = (V, E)$ whose node set $V$ is partitioned into two nonempty sets $V_1$ and $V_2$ in such a way that every edge $E$ joins a node in $V_1$ to a node in $V_2$ and is labeled with a weight. In our projection application, the two partitions are the sets of linguistic units $U_s$ and $U_t$, in the source and target sentence, respectively. We assume that $G$ is complete, that is, each source node is connected to all target nodes and vice versa.[9] Edge weights model the (dis-)similarity between a pair of source and target units.

The optimization problem from Equation (6) identifies the alignment that maximizes the product of link similarities which are equivalent to edges in our bipartite graph. Finding an optimal alignment amounts to identifying a *minimum-weight subgraph* (Cormen, Leiserson, & Rivest, 1990) — a subgraph $G'$ of $G$ that satisfies certain structural constraints (see the discussion below) while incurring a minimal edge cost:

$$\hat{A} = \operatorname*{argmin}_{A \in \mathcal{A}} \sum_{(u_s, u_t) \in A} weight(u_s, u_t) \tag{7}$$

The minimization problem in Equation (7) is equivalent to the maximization problem in (6) when setting $weight(u_s, u_t)$ to:

$$weight(u_s, u_t) = -\log sim(u_s, u_t) \tag{8}$$

---

9. Unwanted alignments can be excluded explicitly by setting their similarity to zero.

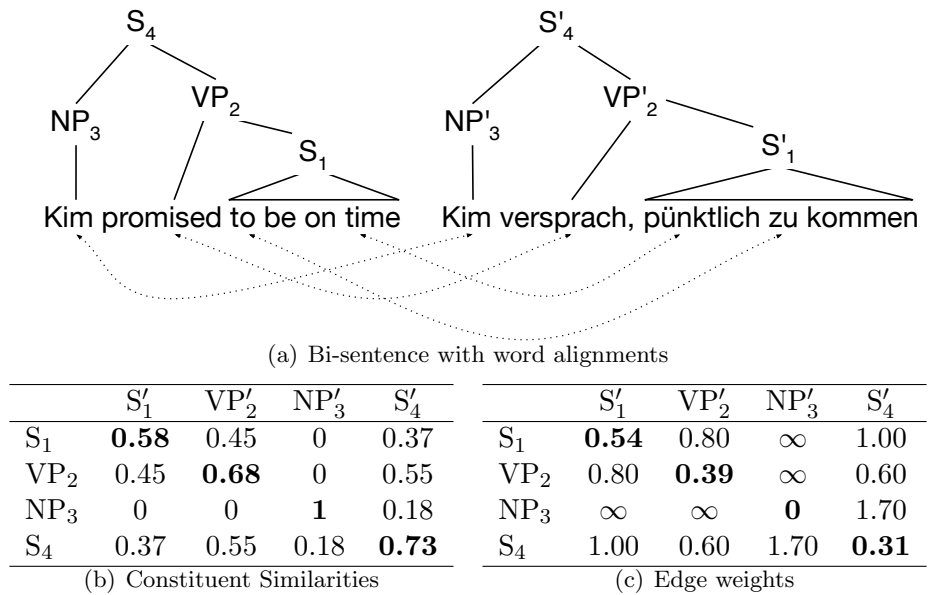

(a) Bi-sentence with word alignments

|         | S'$_1$ | VP'$_2$ | NP'$_3$ | S'$_4$ |         | S'$_1$ | VP'$_2$ | NP'$_3$ | S'$_4$ |
|---------|--------|---------|---------|--------|---------|--------|---------|---------|--------|
| S$_1$   | **0.58** | 0.45  | 0       | 0.37   | S$_1$   | **0.54** | 0.80  | ∞       | 1.00   |
| VP$_2$  | 0.45   | **0.68** | 0     | 0.55   | VP$_2$  | 0.80   | **0.39** | ∞     | 0.60   |
| NP$_3$  | 0      | 0       | **1**   | 0.18   | NP$_3$  | ∞      | ∞       | **0**   | 1.70   |
| S$_4$   | 0.37   | 0.55    | 0.18    | **0.73** | S$_4$ | 1.00   | 0.60    | 1.70    | **0.31** |

(b) Constituent Similarities  (c) Edge weights

Figure 2: Example of bi-sentence represented as an edge weight matrix

As an example, consider Figure 2. It shows the bi-sentence from Figure 1 and its representation as an edge weight matrix for the corresponding complete bipartite graph. The nodes of the graph (S$_1$–S$_4$ for source side and S'$_1$–S'$_4$ for target side) model the sentential constituents. The numbers in Figure 2b are similarity scores, the corresponding edge weights are shown in Figure 2c. High similarity scores correspond to low edge weights. Edges with zero similarity are set to infinity (in practice, to a very large number). Finally, notice that alignments with high similarity scores (low edge weights) occur in the diagonal of the matrix.

In order to obtain complete projection models we must (a) specify the linguistic units over which alignment takes place; (b) define an appropriate similarity function; and (c) formulate the alignment constraints. In the following, we describe two models, one that uses words as linguistic units and one that uses constituents. We also present appropriate similarity functions for these models and detail our alignment constraints.

## 3.2 Word-based Projection

In our first model the linguistic units are *word tokens*. Source and target sentences are represented by sets of words, $U_s = \{w_s^1, w_s^2, \dots\}$ and $U_t = \{w_t^1, w_t^2, \dots\}$, respectively. Semantic alignments here are links between individual words. We can thus conveniently interpret off-the-shelf word alignments as semantic alignments. Formally, this is achieved with the following binary similarity function, which trivially turns a word alignment into an optimal semantic alignment.

$$sim(w_s, w_t) = \begin{cases} 1 & \text{if } w_s \text{ and } w_t \text{ are word-aligned} \\ 0 & \text{else} \end{cases} \qquad (9)$$

Constraints on admissible alignments are often encoded in word alignment models either heuristically (e.g., by enforcing one-to-one alignments as in Melamed, 2000) or by virtue of the translation model used for their computation. For example, the IBM models introduced in the seminal work of Brown, Pietra, Pietra, and Mercer (1993) require each target word to be aligned to exactly one source word (which may be the empty word), and therefore allow one-to-many alignments in one direction. Our experiments use the alignments induced by the publicly available GIZA++ software (Och & Ney, 2003). GIZA++ yields alignments by interfacing the IBM models 1–4 (Brown et al., 1993) with HMM extensions of models 1 and 2 (Vogel, Ney, & Tillmann, 1996). This particular configuration has been shown to outperform several heuristic and statistical alignment models (Och & Ney, 2003). We thus take advantage of the alignment constraints already encoded in GIZA++ and assume that the optimal semantic alignment is given by the set of GIZA++ links. The resulting target language labeling function is:

$$a_t^w(r) = \{w_t \,\|\, \exists\, w_s \in a_s(r) : w_s \text{ and } w_t \text{ are GIZA++ word-aligned}\} \tag{10}$$

This labeling function corresponds to the (implicit) labeling functions employed in other word-based annotation projection models. Such models can be easily derived for different language pairs without recourse to any corpus-external resources. Unfortunately, as discussed in Section 2, automatically induced alignments are often noisy, thus leading to projection errors. Cases in point are function words (e.g., prepositions) and multi-word expressions, which are systematically misaligned due to their high degree of cross-lingual variation.

### 3.3 Constituent-based Projection

In our second model the linguistic units are *constituents*. Source and target sentences are thus represented by constituent sets ($U_s = \{c_s^1, c_s^2, \dots\}$) and ($U_t = \{c_t^1, c_t^2, \dots\}$). A constituent-based similarity function should capture the extent to which $c_s$ and its projection $c_t$ express the same semantic content. We approximate this by measuring *word alignment-based word overlap* between $c_s$ and $c_t$ with Jaccard's coefficient.

Let $yield(c)$ denote the set of words in the yield of a constituent $c$, and $al(c)$ the set of words in the target language aligned to the yield of $c$. Then the word overlap $o$ between a source constituent $c_s$ and a target constituent $c_t$ is defined as:

$$o(c_s, c_t) = \frac{|al(c_s) \cap yield(c_t)|}{|al(c_s) \cup yield(c_t)|} \tag{11}$$

Jaccard's coefficient is asymmetric: it will consider how well the projection of a source constituent $al(c_s)$ matches the target constituent $c_t$, but not vice versa. In order to take target-source and source-target correspondences into account, we measure word overlap in both directions and use their mean as a measure of similarity:

$$sim(c_s, c_t) = (o(c_s, c_t) + o(c_t, c_s))/2 \tag{12}$$

In addition to a similarity measure, a constituent-based projection model must also specify the constraints that characterize the set of admissible alignments $\mathcal{A}$. In this paper, we consider three types of alignment constraints that affect the *number of alignment links per constituent* (in graph-theoretic terms, the *degree of the nodes in $U_s$*). This focus is motivated by patterns that we observe on our gold standard corpus (cf. Section 2). For each English and German constituent, we determined whether it corresponded to none, exactly one, or several constituents in the other language, according to a gold standard word alignment. The majority of constituents correspond to exactly one constituent (67%), followed by a substantial number of one-to-many/many-to-one correspondences (32%), while cases where constituents are not translated (i.e., do not have a corresponding node on the other side of the bi-sentence) are very rare (1%).

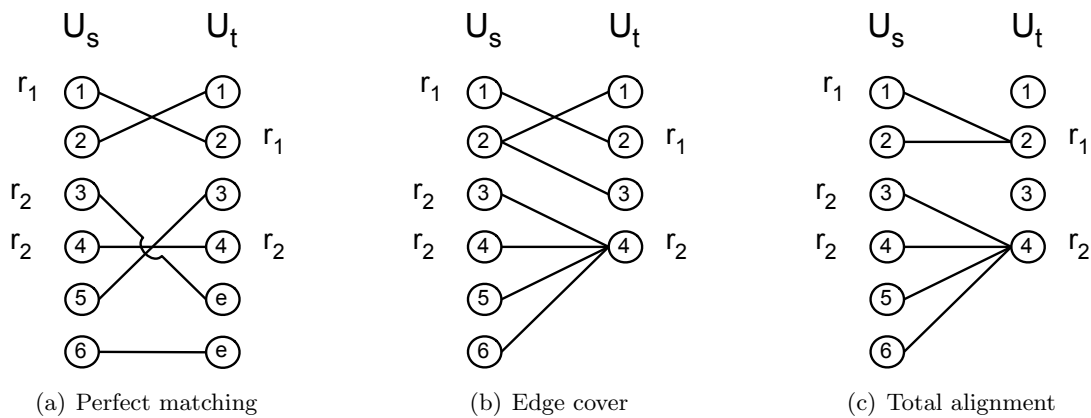

(a) Perfect matching      (b) Edge cover      (c) Total alignment

Figure 3: Constituent alignments and role projections resulting from three families of alignment constraints ($U_s$, $U_t$: source and target constituents; $r_1$, $r_2$: semantic roles).

This analysis indicates that on perfect data, we should expect the best performance from a model that allows one-to-many alignments. However, it is a common finding in machine learning that more restrictive models, even though not appropriate for the data at hand, yield better results by limiting the hypothesis space. In this spirit, we compare three families of admissible alignments which range from the more restrictive to the more permissive, and evaluate them against each other.

### 3.3.1 Alignments as Perfect Matchings

We first consider the most restrictive case, where each constituent has exactly one adjacent alignment edge. Semantic alignments with this property can be thought of as bijective functions: each source constituent is mapped to one target constituent, and vice versa. The resulting bipartite graphs are *perfect matchings*. An example of a perfect bipartite matching is given in Figure 3a. Note that the target side contains two nodes labeled (e), a shorthand for "empty" node. Since bi-sentences will often differ in size, the resulting graph partitions will have different sizes as well. In such cases, we introduce empty nodes in the smaller partition to enable perfect matching. Empty nodes are assigned a similarity of zero with all other nodes. Alignments to empty nodes (such as for source nodes (3) and (6)) are ignored for the purposes of projection; this gives the model the possibility of abstaining from aligning a node when no good correspondence is found.

Perfect matchings assume a strong equivalence between the constituent structures of the two languages; neither of the alignments in Figure 3(b) and 3(c) is a perfect matching.

Perfect matchings cannot model one-to-many matches, i.e., cases where semantic material expressed by one constituent in one language is split into two constituents in the other language. This means that perfect matchings are most appropriate when the source and target role annotations span exactly one constituent. While this is not always the case, perfect matchings also have an advantage over more expressive models: by allowing each node only one adjacent edge, they introduce strong competition between edges. As a result, errors in the word alignment can be corrected to some extent.

Perfect matchings can be computed efficiently using algorithms for network optimization (Fredman & Tarjan, 1987) in time approximately cubic in the total number of constituents in a bi-sentence ($O(|U_s|^2 \log |U_s| + |U_s|^2|U_t|)$). Furthermore, perfect matchings are equivalent to the well-known *linear assignment problem*, for which many solution algorithms have been developed (e.g., Jonker and Volgenant 1987, time $O(\max(|U_s|, |U_t|)^3)$).

### 3.3.2 ALIGNMENTS AS EDGE COVERS

We next consider *edge covers*, a generalization of perfect matchings. Edge covers are bipartite graphs where each source and target constituent is adjacent to *at least* one edge. This is illustrated in Figure 3b, where all source and target nodes have one adjacent edge (i.e., alignment link), and some nodes more than one (see source node (2) and target node (4)). Edge covers impose weaker correspondence assumptions than perfect matchings, since they allow one-to-many alignments between constituents in either direction.[10] So, in theory, edge covers have a higher chance of delivering a correct role projection than perfect matchings when the syntactic structures of the source and target sentences are different. They can also deal better with situations where semantic roles are assigned to more than one constituent in one of the languages (cf. source nodes (3) and (4), labeled with role $r_2$, in the example graph). Notice that perfect matchings as shown in Figure 3a are also edge covers, whereas the graph in Figure 3c is not, as the target-side nodes (1) and (3) have no adjacent edges.

Eiter and Mannila (1997) develop an algorithm for computing optimal edge covers. They show that minimum-weight edge covers can be reduced to minimum weight perfect matchings (see above) of an auxiliary bipartite graph with two partitions of size $|U_s| + |U_t|$. This allows the computation of minimum weight edge covers in time $O((|U_s| + |U_t|)^3) = O(max(|U_s|, |U_t|)^3)$, which is also cubic in the number of constituents in the bi-sentence.

### 3.3.3 TOTAL ALIGNMENTS

The last family of admissible alignments we consider are total alignments. Here, each source constituent is linked to some target constituent (i.e.,the alignment forms a total function on the source nodes). Total alignments do not impose any constraints on the target nodes, which can therefore be linked to an arbitrary number of source nodes, including none. Total alignments are the most permissive alignment class. In contrast to perfect matchings and edge covers, constituents in the target sentence can be left unaligned. Total alignments can

---

10. The general definition of edge covers also allows many-to-many alignments. However, *optimal* edge covers according to Equation (7) cannot be many-to-many, since the weight of edge covers with many-to-many alignments can never be minimal: From each many-to-many edge cover, one edge can be removed, resulting in an edge cover with a lower weight.

be computed by linking each source node to its maximally similar target node:

$$\hat{A}^t = \{(c_s, c_t) \mid c_s \in U_s \wedge c_t = \operatorname*{argmax}_{c_t' \in U_t} sim(c_s, c_t')\} \tag{13}$$

Due to the independence of source nodes, this local optimization results in a globally optimal alignment. The time complexity of this procedure is quadratic in the number of constituents, $O(|U_s||U_t|)$.

An example is shown in Figure 3c, where source constituents (1) and (2) correspond to target constituent (2), and source constituents (3)–(6) correspond to (4). Target constituents (1) and (3) are not aligned. The quality of total alignments relies heavily on the underlying word alignment. Since there is little competition between edges, there is a tendency to form alignments mostly with high (similarity) scoring target constituents. In practice, this means that potentially important, but idiosyncratic, target constituents with low similarity scores will be left unaligned.

## 3.4 Noise Reduction

As discussed in Section 2, noise elimination techniques are an integral part of projection architectures. Although our constituent-based model will not be overly sensitive to noise — we expect the syntactic information to compensate for alignment errors — the word-based model will be more error-prone since it relies solely on automatically obtained alignments for transferring semantic roles. Below we introduce three filtering techniques that either correct or discard erroneous projections.

### 3.4.1 FILLING-THE-GAPS

According to our definition of projection in Equation (5), the span of a projected role $r$ corresponds to the union of all target units that are aligned to source units labeled with $r$. This definition is sufficient for constituent-based projection models, where roles rarely span more than one constituent but will yield many wrong alignments for word-based models where roles will typically span several source units (i.e., words). Due to errors and gaps in the word alignment, the target span of a role will often be a non-contiguous set of words. We can repair non-contiguous projections where the first and last word has been projected correctly by applying a simple heuristic which fills the gaps in a target role span. Let $pos$ be the index of a word token $t$ in a given sentence, and by extension the set of indices for a set of words. The target span of role $r$ without any gaps is then defined as:

$$a_t^{cc}(r) = \{u \mid \min(pos(a_t(r))) \leq pos(u) \leq \max(pos(a_t(r)))\} \tag{14}$$

We apply this heuristic to all word-based models in this article.

### 3.4.2 WORD FILTERING

This technique removes words form a bi-sentence prior to projection according to certain criteria. We apply two intuitive instantiations of word filtering in our experiments. The first removes non-content words, i.e., all words which are not adjectives, adverbs, verbs, or nouns, from the source and target sentences, since alignments for non-content words are notoriously unreliable and may negatively impact the similarity computations. The second filter removes all words which remain unaligned in the output of the automatic word alignment. Both filters aim at distinguishing genuine word alignments from noisy ones and speed up the computation of semantic alignments.

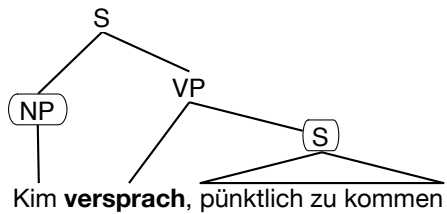

Figure 4: Argument filtering (predicate in boldface, potential arguments in boxes).

### 3.4.3 ARGUMENT FILTERING

Our last filter applies only to constituent-based models defined over full parse trees. Previous work in shallow semantic parsing has demonstrated that not all nodes in a tree are equally probable as semantic roles for a given predicate (Xue & Palmer, 2004). In fact, assuming a perfect parse, there is a "set of likely arguments", to which almost all semantic roles should be assigned. This set of likely arguments consists of all constituents which are a child of some ancestor of the predicate, provided that (a) they do not dominate the predicate themselves and (b) there is no sentence boundary between a constituent and its predicate. This definition covers long-distance dependencies such as control constructions for verbs, or support constructions for nouns, and can be extended to accommodate coordination. We apply this filter to reduce the size of the *target* tree. In the example in Figure 4, all tree nodes are removed except the NP *Kim* and the S *pünktlich zu kommen*.

### 3.5 Discussion

We have presented a framework for the bilingual projection of semantic roles which is based on the notion of semantic alignment. We have discussed two broad instantiations of the framework, namely word-based and constituent-based models. In the latter case, we operationalize the search for an optimal semantic alignment as a graph optimization problem. Specifically, each bi-sentence is conceptualized as a bipartite graph. Its nodes correspond to the syntactic constituents of the bi-sentence, and its weighted edges to the cross-lingual pairwise similarity between constituents. Assumptions about the semantic correspondence between the languages are formalized as constraints on the graph structure.

We have also discussed three families of constraints. Perfect matching forces correspondences between English and German constituents to be bijective. In contrast, total alignments assume a looser correspondence by leaving constituents on the target side unaligned. Edge covers occupy a middle ground, assuming that all constituents must be aligned without strictly enforcing one-to-one alignments. While perfect matching is linguistically implausible, by assuming no structural divergence between languages, it can overcome word alignment errors. Total alignments can model structural changes and are therefore linguis-tically more appropriate, but at the same time more sensitive to alignment errors. Finding an optimal alignment corresponds to finding the optimal subgraph consistent with our constraints. Efficient algorithms exist for this problem. Finally, we have introduced a small number of filtering techniques which further reduce the impact of alignment errors.

| LingUnit | Similarity | Correspondence | BiGraph | Complexity |
|----------|-----------|----------------|---------|-----------|
| words | binary | one-to-one | n/a | linear |
| constituents | overlap | one-to-one | perfect matching | cubic |
| constituents | overlap | one-to-at-least-one | edge cover | cubic |
| constituents | overlap | one-to-many | total | quadratic |

Table 4: Framework instantiations

Our models and their properties are summarized in Table 4. They vary along the following dimensions: the linguistic units employed (LingUnit), the similarity measure (Similarity), their assumptions about semantic correspondence (Correspondence) and the structure of the bipartite graph this entails (BiGraph). We also mention the complexity of their computation (Complexity). We empirically assess their performance in the following sections.

## 4. Experiments

We now describe our evaluation of the framework developed in Section 3. We present two experiments, both of which consider the projection of semantic roles from English sentences onto their German translations, and evaluate them against German gold standard role annotation. Experiment 1 uses gold standard data for both syntactic and semantic annotation. This "oracle setting" assesses the potential of role projection on its own, separating the errors due to translational divergence and our modeling assumptions from those incurred by preprocessing (e.g., parsing and automatic alignment). Experiment 2 investigates a more practical setting which employs automatic tools for syntactic and semantic parsing, thus approximating the conditions of large-scale role projection on parallel corpora.

### 4.1 Setup

#### 4.1.1 DATA

All models were evaluated on the parallel corpus described in Section 2. The corpus was randomly shuffled and split into a development and a test set (each 50% of the data). Table 5 reports the number of tokens, sentences, frames, and arguments in the development and test set for English and German.

Word alignments were computed with the GIZA++ toolkit (Och & Ney, 2003). We used the entire English-German Europarl bitext as training data to induce alignments for both directions (source-target, target-source), with the default GIZA++ settings. Following common practice in Machine Translation, the alignments were symmetrized using the intersection heuristic (Koehn, Och, & Marcu, 2003), which is known to lead to high-precision alignments. We also produced manual word alignments for all sentences in our corpus, using the GIZA++ alignments as a starting point and following the Blinker annotation guidelines (Melamed, 1998).

| Language | Tokens | Sentences | Frames | Roles |
|----------|--------|-----------|--------|-------|
| Dev-EN   | 11,585 | 491       | 491    | 2,423 |
| Test-EN  | 12,019 | 496       | 496    | 2,465 |
| Dev-DE   | 11,229 | 491       | 491    | 2,576 |
| Test-DE  | 11,548 | 496       | 496    | 2,747 |

Table 5: Statistics of gold standard parallel corpus broken down into development (Dev) and test (Test) set.

### 4.1.2 Method

We implemented the four models shown in Table 4 on their own and with the filtering techniques introduced in Section 3.4. This resulted in a total of sixteen models, all of which were evaluated on the development set. Results for the best-performing models were next validated on the test set. We found the practical runtime of our experiment to be dominated by input/output XML processing rather than the optimization problem itself.

In both experiments, constituent-based models were compared against the word-based model, which we treat as a baseline. This latter model is relatively simple: projection relies exclusively on word alignments, it does not require syntactic analysis, and has linear time complexity. It thus represents a good point of comparison for models that take linguistic knowledge into account.

### 4.1.3 Evaluation Measure

We measure model performance using labeled Precision, Recall, and F1 in the "Exact Match" condition, i.e., both the label and the span of the projected English role have to match the German gold standard role to count as a true positive. We also assess whether differences in performance are statistically significant using stratified shuffling (Noreen, 1989), an instance of assumption-free approximate randomization testing (see Yeh, 2000, for a discussion).[11] Whenever we discuss changes in F1, we refer to absolute (rather than relative) differences.

### 4.1.4 Upper Bound

Our annotation study (see Table 2, Section 2.2) obtained an inter-annotator agreement of 0.84 in the Span Match condition (annotation of the same roles with the same span). This number can be seen as a reasonable upper bound for the performance of an automatic semantic role labeling system *within* a language. It is more difficult to determine a ceiling for the projection task, since in addition to inter-annotator agreement, we have to take into account the effect of bilingual divergence. Our annotation study did provide an estimate of the former, but not of the latter. In default of a method for measuring bilingual agreement on spans, we adopt the monolingual Span Match agreement as an upper bound for our projection experiments. Note, however, that this upper bound is not strict — a system with an oracle should be able to outperform it.

---

11. The software is available at `http://www.nlpado.de/~sebastian/sigf.html`.

| No Filter | | | | NA Filter | | | |
|---|---|---|---|---|---|---|---|
| Model | Prec | Rec | F1 | Model | Prec | Rec | F1 |
| WordBL | 52.0 | 46.2 | 48.9 | WordBL | 52.0 | 46.2 | 48.9 |
| PerfMatch | 75.8 | 57.1 | 65.1 | **PerfMatch** | **81.4** | **69.4** | **74.9** |
| **EdgeCover** | **71.7** | **61.8** | **66.4** | EdgeCover | 77.9 | 69.3 | 73.3 |
| Total | 68.9 | 61.3 | 64.9 | Total | 78.8 | 70.0 | 74.1 |
| UpperBnd | 85.0 | 83.0 | 84.0 | UpperBnd | 85.0 | 83.0 | 84.0 |

| NC Filter | | | | Arg Filter | | | |
|---|---|---|---|---|---|---|---|
| Model | Prec | Rec | F1 | Model | Prec | Rec | F1 |
| WordBL | 37.1 | 32.0 | 34.4 | WordBL | 52.0 | 46.2 | 48.9 |
| **PerfMatch** | **79.4** | **62.2** | **69.8** | PerfMatch | 88.8 | 56.2 | 68.8 |
| EdgeCover | 75.0 | 63.0 | 68.5 | **EdgeCover** | **81.4** | **69.7** | **75.1** |
| Total | 69.7 | 60.1 | 64.5 | Total | 81.2 | 69.6 | 75.0 |
| UpperBnd | 85.0 | 83.0 | 84.0 | UpperBnd | 85.0 | 83.0 | 84.0 |

Table 6: Model comparison on the development set using gold standard parses and semantic roles and four filtering techniques: no filtering (No Filter), removal of non-aligned words (NA Filter), removal of non-content words (NC Filter), and removal of non-arguments (Arg Filter). Best performing models are indicated in boldface.

## 4.2 Experiment 1: Projection on Gold Standard Data

In Experiment 1, we use manually annotated semantic roles and hand-corrected syntactic analyses for the constituent-based projection models. As explained in Section 4.1, we first discuss our results on the development set. The best model instantiations are next evaluated on the test set.

### 4.2.1 DEVELOPMENT SET

Table 6 shows the performance of our models on their own (No Filter) and in combination with filtering techniques. In the No Filter condition, the word-based model (WordBL) yields a modest F1 of 48.9% with the application of filling-the-gaps heuristic[12] (see Section 3.4 for details). In the same condition, constituent-based models deliver an F1 increase of approximately 20% (all differences between WordBL and constituent-based models are significant, $p < 0.01$). The EdgeCover model performs significantly better than total alignments (Total, $p < 0.05$) but comparably to perfect matchings (PerfMatch).

Filtering schemes generally improve the resulting projections for the constituent-based models. When non-aligned words are removed (NA Filter), F1 increases by 9.8% for Perf-Match, 6.9% for EdgeCover, and 9.2% for Total. PerfMatch and Total are the best performing models with the NA Filter. They significantly outperform EdgeCover ($p < 0.05$) in the same condition and all constituent-based models in the No Filter condition ($p < 0.01$). The word-based model's performance remains constant. By definition WordBL considers aligned words only; thus, the NA Filter has no impact on its performance.

---

12. Without filling-the-gaps, F1 drops to 40.8%.

Moderate improvements are observed for constituent-based models when non-content words are removed (NC filter). PerfMatch performs best in this condition. It is significantly better than PerfMatch, EdgeCover or Total in the No Filter condition ($p < 0.01$), but significantly worse than all constituent-based models in the NA filter condition ($p < 0.01$). The NC filter yields significantly worse results for WordBL ($p < 0.01$). This is not surprising, since the word-based model cannot recover words that have been deleted by a filter, such as role-initial prepositions or subordinating conjunctions.

Note also that combinations of different filtering techniques can be applied to our constituent- and word-based models. For example, we can create a constituent-based model where non-aligned and content words are removed as well as non-arguments. For the sake of brevity, we do not present results for filter combinations here as they generally do not improve results further. We find that combining filters tends to remove a large number of words, and as a result, good alignments are also removed.

Overall, we obtain best performing models when non-argument words are removed (Arg Filter). Arg Filter is an aggressive filtering technique, since it removes all constituents but those likely to occupy argument positions. EdgeCover and Total are significantly better than PerfMatch in the Arg Filter condition ($p < 0.01$), but perform comparably to PerfMatch when the NA Filter is applied. Moreover, EdgeCover and Total construct almost identical alignments. This indicates that the two latter models behave similarly when the alignment space is reduced after removing many possible bad alignments, despite imposing different constraints on the structure of the bipartite graph. Interestingly, the strict correspondence constraints imposed by PerfMatch result in substantially different alignments. Recall that PerfMatch attempts to construct a bilingual one-to-one mapping between arguments. When no direct correspondence can be identified for source nodes, it abstains from projecting. As a result, the alignment produced by PerfMatch shows the highest precision among all models (88.8%), which is offset by the lowest recall (56.2%). These results tie in with our earlier analysis of constituent alignments (Section 3.3), where we found that about one-third of the corpus correspondences are of the one-to-many type. Consider the following example:

(15)    The Charter means [$_{NP}$ an opportunity to bring the EU closer to the people].

　　　　Die Charta bedeutet [$_{NP}$ eine Chance], [$_{S}$ die EU den　Bürgern näherzubringen].
　　　　The Charter means　　[$_{NP}$ a　　chance], [$_{S}$ the EU to the citizens to bring closer].

Ideally, the English NP should be aligned to both the German NP and S. EdgeCover, which can model one-to-many relationships, acts "confidently" and aligns the NP to the German S to maximize the overlap similarity, incurring both a precision and recall error. PerfMatch, on the other hand, cannot handle one-to-many alignments and acts "cautiously" and makes only a recall error by aligning the English NP to an empty node. Thus, according to our evaluation criteria, the analysis of EdgeCover is deemed worse than that of PerfMatch, even though the former is partly correct.

In sum, filtering improves the resulting projections by making the syntactic analyses of the source and target sentences more similar to each other. Best results are observed in NA Filter (PerfMatch) and Arg Filter conditions (Total and EdgeCover). Finally, note that the best models obtain precision figures that are close to or above the upper bound. The best recall values are around 70%, significantly below the upper bound of 83%. Aside from wrongly assigned roles, recall errors are due to short semantic roles (e.g., pronouns), for which the intersective word alignment often does not contain any alignment links, so that projection cannot proceed.

| Intersective word alignment | | | | Manual word alignment | | | |
|---|---|---|---|---|---|---|---|
| Model | Prec | Rec | F1 | Model | Prec | Rec | F1 |
| WordBL | 52.9 | 47.4 | 50.0 | WordBL | 76.1 | 73.9 | 75.0 |
| **EdgeCover** | **86.6** | **75.2** | **80.5** | **EdgeCover** | **86.0** | **81.8** | **83.8** |
| PerfMatch | 85.1 | 73.3 | 78.8 | PerfMatch | 82.8 | 76.3 | 79.4 |
| UpperBnd | 85.0 | 83.0 | 84.0 | UpperBnd | 85.0 | 83.0 | 84.0 |

Table 7: Model performance on the test set with intersective and manual alignments. Edge-Cover uses Arg Filter and PerfMatch uses NA Filter. Best performing models are indicated in boldface.

### 4.2.2 TEST SET

Our experiments on the test set focus on models which have obtained best results on the development set using a specific filtering technique. In particular, we report performance for EdgeCover and PerfMatch in the Arg Filter and NA Filter conditions, respectively. In addition, we assess the effect of the automatic word alignment on these models by using both intersective and manual word alignments.

Our results are summarized in Table 7. When intersective alignments are used (left-hand side), EdgeCover performs numerically better than PerfMatch, but the difference is not statistically significant. This corresponds to our findings on the development set.[13] The right-hand side shows the results when manually annotated word alignments are used. As can be seen, the performance of WordBL increases sharply from 50.0% to 75.0% (F1). This underlines the reliance of the word-based model on clean word alignments. Despite this performance improvement, WordBL still lags behind the best constituent-based model by approximately 9% F1. This means that there are errors made by the word-based model that can be corrected by constituent-based models, mostly cases where translation introduced material in the target sentence that cannot be word-aligned to any expressions in the source sentence or recovered by the filling-the-gaps heuristic. An example is shown below, where the translation of *clarification* as *more detailed explanation* leads to the introduction of two German words, *die näheren.* These words are unaligned at the word level and thus do not form part of the role when word-based projection is used.

(16)  [Commissioner Barnier's clarification]Role

[*die näheren*  Erläuterungen von Kommissar  Barnier]Role
[*the more detailed* explanations  of  Commissioner Barnier]Role

---

13. Our results on the test set are slightly higher in comparison to the development set. The fluctuation reflects natural randomness in the partitioning of our corpus.

| Evaluation condition | Prec | Rec | F1 |
|---|---|---|---|
| All predicates | 81.3 | 58.6 | 68.1 |
| Verbs only | 81.3 | 63.8 | 71.5 |

Table 8: Evaluation of Giuglea and Moschitti's (2004) shallow semantic parser on the English side of our parallel corpus (test set)

Constituent-based models generally profit less from cleaner word alignments. Their performance increases by 1%–3% F1. The improvement is due to higher recall (approximately 5% in the case of EdgeCover) but not precision. In other words, the main effect of the manually corrected word alignment is to make possible the projection of previously unavailable roles. EdgeCover performs close to the human upper bound when gold standard alignments are used. Under noise-free conditions it is able to account for one-to-many constituent alignments, and thus models our corpus better.

Aside from alignment noise, most of the errors we observe in the output of our models are due to translational divergences, or problematic monolingual role assignments, such as pronominal adverbs in German. Many German verbs such as *glauben* ("believe") exhibit a diathesis alternation: they subcategorize either for a prepositional phrase (*X glaubt an Y*, "X believes in Y"), or for an embedded clause which must be preceded by the pronominal adverb *daran* (*X glaubt daran, dass Y*, "X believes that Y"). Even though the pronominal adverb forms part of the complement clause (and therefore also of the role assigned to it), it has no English counterpart. In contrast to example (16) above, the incomplete span *dass Y* forms a complete constituent. Unless it is removed by Arg Filter prior to alignment, this error cannot be corrected by the use of constituents.

### 4.3 Experiment 2: Projection with Automatic Roles

In this experiment, we evaluate our projection models in a more realistic setting, using automatic tools for syntactic and semantic parsing.

#### 4.3.1 Preprocessing

For this experiment, we use the uncorrected syntactic analyses of the bilingual corpus as provided by Collins' (1997) and Dubey's (2005) parsers (cf. Section 2.1). We automatically assigned semantic roles using a state-of-the-art semantic parser developed by Giuglea and Moschitti (2004). We trained their parser on the FrameNet corpus (release 1.2) using their "standard" features and not the "extended" set which is based on PropBank and would have required a PropBank analysis of the entire FrameNet corpus.

We applied the shallow semantic parser to the English side of our parallel corpus to obtain semantic roles, treating the frames as given.[14] The task involves locating the frame elements in a sentence and finding the correct label for a particular frame element. Table 8 shows an evaluation of the parser's output on our test set against the English gold standard annotation. Giuglea and Moschitti report an accuracy of 85.2% on the role classification task, using the "standard" feature set.[15] Our results are not strictly comparable to theirs, since we identify the role-bearing constituents in addition to assigning them a label. Our performance can thus be expected to be worse, since we inherit errors from the frame element identification stage. Secondly, our test set differs from the training data in vocabulary (affecting the lexical features) and suffers from parsing errors. Since Giuglea and Moschitti's (2004) implementation can handle only verbs, we also assessed performance on the subset of verbal predicates (87.5% of test tokens). The difference between the complete and verbs-only data sets amounts to 3.4% F1, which represents the unassigned roles for nouns.

---

14. This decomposition of frame-semantic parsing has been general practice in recent role labeling tasks, e.g. at Senseval-3 (Mihalcea & Edmonds, 2004).

| Model | Filter | Prec | Rec | F1 |
|---|---|---|---|---|
| WordBL | | 52.5 | 34.5 | 41.6 |
| **PerfMatch** | **NA Filter** | **73.0** | **45.4** | **56.0** |
| EdgeCover | Arg Filter | 70.0 | 45.1 | 54.9 |
| UpperBnd | | 81.3 | 58.6 | 68.1 |

Table 9: Performance of best constituent-based model on the test set (automatic syntactic and semantic analysis, intersective word alignment)

### 4.3.2 SETUP

We report results for the word-based baseline model, and the best projection models from Experiment 1, namely PerfMatch (NA filter) and EdgeCover (Arg Filter). We use the the complete test set (including nouns and adjectives) in order to make our evaluation comparable to Experiment 1.

### 4.3.3 RESULTS

The results are summarized in Table 9. Both PerfMatch (NA Filter) and EdgeCover (Arg Filter) perform comparably at 55–56% F1. This is approximately 25 points F1 worse than the results obtained on manual annotation (compare Table 9 to Table 7). WordBL's performance (now 41.6% F1) degrades less (around 8% F1), since it is only affected by semantic role assignment errors. However, consistently with Experiment 1, the constituent-based models sill outperform WordBL by more than 10% F1 ($p < 0.01$). These results underscore the ability of bracketing information, albeit noisy, to correct and extend word alignment. Although the Arg Filter performed well in Experiment 1, we observe less of an effect here. Recall that the filter uses not only bracketing, but also dominance information, and is therefore particularly vulnerable to misparses. Like in Experiment 1, we find that our models yield overall high precision but low recall. Precision drops by 15% F1 when automatic annotations are used, whereas recall drops by 30%; however, note that this drop includes about 5% nominal roles that fall outside the scope of the shallow semantic parser.

Further analysis showed that parsing errors form a large source of problems in Experiment 2. German verb phrases are particularly problematic. Here, the relatively free word order combines with morphological ambiguity to produce ambiguous structures, since third person plural verb forms (FIN) are identical to infinitival forms (INF). Consider the following English sentence, (17a), and two syntactic analyses of its German translation, (17b)/(17c):

(17)    a.    when we recognize [that we have to work on this issue]

       b.    wenn wir erkennen$_{FIN}$, [dass wir [daran arbeiten$_{INF}$] müssen$_{FIN}$]
          "when we recognize [that we have to [work on it]]"

       c.    wenn wir [erkennen$_{INF}$, [dass wir daran arbeiten$_{FIN}$]] müssen$_{FIN}$
          "when we have to [recognize that [we work on it]]"

---

15. See the work of Giuglea and Moschitti (2006) for an updated version of their shallow semantic parser.

| Band | Prec | Rec | F1 |
|---|---|---|---|
| Error 0 | 85.1 | 74.1 | 79.2 |
| Error 1 | 75.9 | 34.6 | 47.6 |
| Error 2+ | 40.7 | 18.5 | 25.4 |

Table 10: PerfMatch's performance in relation to error rate in the automatic semantic role labeling (Error 0: no labeling errors, Error 1: one labeling error, Error 2+: two or more labeling errors).

(17b) gives the correct syntactic analysis, but the parser we used produced the highly implausible (17c). As a result, the English sentential complement *that we have to work on this issue* cannot be aligned to a single German constituent, nor to a combination of them. In this situation, PerfMatch will generally not align the constituent at all and thus sacrifice recall. EdgeCover (and Total) will produce a (wrong) alignment and sacrifice precision.

Finally, we evaluated the impact of semantic role labeling errors on projection. We split the semantic parser's output into three bands: (a) sentences with no role labeling errors (Error 0, 35% of the test set), (b) sentences with one labeling error (Error 1, 33% of the test set), and (c) sentences with two or more labeling errors (Error 2+, 31% of the test set). Table 10 shows the performance of the best model, PerfMatch (NA filter), for each of these bands. As can be seen, when the output of the automatic labeler is error-free, projection attains an F1 of 79.2%, comparable to the results obtained in Experiment 1.[16] Even though projection clearly degrades with the quality of the semantic role input, PerfMatch still delivers projections with high precision for the Error 1 band. As discussed above, the low recall values for bands Error 1 and 2+ result from the labeler's low recall.

## 5. Related Work

Previous work on annotation projection has primarily focused on annotations spanning short linguistic units. These range from POS tags (Yarowsky & Ngai, 2001), to NP chunks (Yarowsky & Ngai, 2001), dependencies (Hwa et al., 2002), and word senses (Bentivogli & Pianta, 2005). A different strategy for the cross-lingual induction of frame-semantic information is presented by Fung and Chen (2004), who do not require a parallel corpus. Instead, they use a bilingual dictionary to construct a mapping between FrameNet entries and concepts in HowNet, an on-line ontology for Chinese.[17] In a second step, they use HowNet knowledge to identify monolingual Chinese sentences with predicates that instantiate these concepts, and label their arguments with FrameNet roles. Fung and Chen report high accuracy values, but their method relies on the existence of resources which are presumably unavailable for most languages (in particular, a rich ontology). Recently, Basili et al. (2009) propose an approach to semantic role projection that is not word-based and does not employ syntactic information. Using a phrase-based SMT system, they heuristically assemble target role spans out of phrase translations, defining phrase similarity in terms of translation probability. Their method occupies a middle ground between word-based projection and constituent-based projection.

---

16. It is reasonable to assume that in these sentences, at least the relevant part of the syntactic analysis is correct.

The work described in this article relies on a parallel corpus for harnessing information about semantic correspondences. Projection works by creating semantic alignments between constituents. The latter correspond to nodes in a bipartite graph, and the search for the best alignment is cast as an optimization problem. The view of computing optimal alignments by graph matching is relatively widespread in the machine translation literature (Melamed, 2000; Matusov, Zens, & Ney, 2004; Tiedemann, 2003; Taskar, Lacoste-Julien, & Klein, 2005). Despite individual differences, most approaches formalize word alignment as a minimum-weight matching problem, where each pair of words in a bi-sentence is associated with a score representing the desirability of that pair. The alignment for the bi-sentence is the highest scoring matching under some constraints, for example that matchings must be one-to-one. Our work applies graph matching to the level of constituents and compares a larger class of constraints (see Section 3.3) than previous approaches. For example, Taskar et al. (2005) examine solely perfect matchings and Matusov et al. (2004) only edge covers.

A number of studies have addressed the constituent alignment problem in the context of extracting of translation patterns (Kaji, Kida, & Morimoto, 1992; Imamura, 2001). However, most approaches only search for pairs of constituents which are perfectly word aligned, an infeasible strategy when alignments are obtained automatically. Other work focuses on the constituent alignment problem, but uses greedy search techniques that are not guaranteed to find an optimal solution (Matsumoto, Ishimoto, & Utsuro, 1993; Yamamoto & Matsumoto, 2000). Meyers, Yangarber, and Grishman (1996) propose an algorithm for aligning parse trees which is only applicable to isomorphic structures. Unfortunately, this restriction limits their application to structurally similar languages and high-quality parse trees.

Although we evaluate our models only on the semantic role projection task, we believe they also show promise in the context of SMT, especially for systems that use syntactic information to enhance translation quality. For example, Xia and McCord (2004) exploit constituent alignment for rearranging sentences in the source language so as to make their word order similar to that of the target language. They learn tree reordering rules by aligning constituents heuristically, using an optimization procedure analogous to the total alignment model presented in this article. A similar approach is described in a paper by Collins, Koehn, and Kučerová (2005); however, the rules are manually specified and the constituent alignment step reduces to inspection of the source-target sentence pairs. The different alignment models presented in this article could be easily employed for the reordering task common to both approaches.

---

17. For information on HowNet, see `http://www.keenage.com/zhiwang/e_zhiwang.html`.

## 6. Conclusions

In this article, we have argued that parallel corpora show promise in relieving the lexical acquisition bottleneck for new languages. We have proposed annotation projection as a means of obtaining FrameNet annotations automatically, using resources available for English and exploiting parallel corpora. We have presented a general framework for projecting semantic roles that capitalizes on the use of constituent information during projection, and modelled the computation of a constituent alignment as the search for an optimal subgraph in a bipartite graph. This formalization allows us to solve the search problem efficiently using well-known graph optimization methods. Our experiments have focused on three modeling aspects: the level of noise in the linguistic annotation, constraints on alignments, and noise reduction techniques.

We have found that constituent information yields substantial improvements over word alignments. Word-based models offer a starting point for low-density languages for which parsers are not available. However, word alignments are too noisy and fragmentary to deliver accurate projections for annotations with long spans such as semantic roles. Our experiments have compared and contrasted three constituent-based models which differ in their assumptions regarding cross-lingual correspondence (total alignments, edge covers, and perfect matchings). Perfect matchings, a restrictive alignment model that enforces one-to-one alignments, performed most reliably across all experimental conditions. In particular, its precision surpassed all other models. This indicates that a strong semantic correspondence can be assumed as a modelling strategy, at least for English and German and the parsing tools available for these languages. As a side effect, the performance of constituent-based models increases only slightly when manual word alignments are used, which means that near-optimal results can be obtained using automatic alignments.

As far as alignment noise reduction techniques are concerned, we find that removing non-aligned words (NA Filter) and non-arguments (Arg Filter) yields the best results. Both filters are independent of the language pair and make only weak assumptions about the underlying linguistic representations in question. The choice of the best filter depends on the goals of projection. Removing non-aligned words is relatively conservative and tends to balance precision and recall. In contrast, the more aggressive filtering of non-arguments yields projections with high precision and low recall. Arguably, for training shallow semantic parsers on the target language (Johansson & Nugues, 2006), it is more desirable to have access to high-quality projections. However, there is a number of options for increasing precision subsequent to projection that we have not explored in this article. One fully automatic possibility is generalization over multiple occurrences of the same predicate to detect and remove suspicious projection instances, e.g. following the work by Dickinson and Lee (2008). Another direction is postprocessing by annotators, e.g., by adopting the "annotate automatically, correct manually" methodology used to provide high volume annotation in the Penn Treebank project. Our models could also be used in a semi-supervised setting, e.g., to provide training data for unknown predicates.

The extensions and improvements of the framework presented here are many and varied. Firstly, we believe that the models developed in this article are useful for other semantic role paradigms besides FrameNet, or indeed for other types of semantic annotation. Potential applications include the projection of PropBank roles[18], discourse structure, and named entities. As mentioned earlier, our models are also relevant for machine translation and could be used for the reordering of constituents. Our results indicate that syntactic knowledge on the target side plays an important role in projecting annotations with longer spans. Unfortunately, for many languages there are no broad-coverage parsers available. However, it may not be necessary to obtain complete parses for the semantic role projection task. Two types of syntactic information are especially valuable here: bracketing information (which guides projection towards linguistically plausible role spans) and knowledge about the arguments of sentence predicates. Bracketing information can be acquired in an unsupervised fashion (Geertzen, 2003). Argument structure information could be obtained from dependency parsers (e.g., McDonald, 2006) or partial parsers that are able to identify predicate-argument relations (e.g., Hacioglu, 2004). Another interesting direction concerns the combination of longer phrases, like those provided by phrase-based SMT systems, with constituent information obtained from the output of a parser or chunker.

The experiments presented in this article made use of a simple semantic similarity measure based on word alignment. A more sophisticated approach could have combined the alignment scores with information provided in a bilingual dictionary. Inspired by cross-lingual information retrieval, Widdows, Dorow, and Chan (2002) propose a bilingual vector model. The underlying assumption is that words that have similar co-occurrences in a parallel corpus are also semantically similar. Source and target words are represented as $n$-dimensional vectors whose components correspond to the most frequent content words in the source language. In this framework, the similarity of any source-target word pair can be computed using a geometric measure such as cosine or Euclidean distance. The more recent polylingual topic models (Mimno, Wallach, Naradowsky, Smith, & McCallum, 2009) offer a probabilistic interpretation of a similar idea.

In this article, we have limited ourselves to parallel sentences where the frame is preserved. This allows us to transfer roles directly from the source onto the target language without having to acquire knowledge about possible translational divergences first. A generalization of the framework presented here could adopt a strategy where some form of *mapping* is applied during projection, akin to the transfer rules used in machine translation. Thus far, we have only explored models applying the *identity* mapping. Knowledge about other possible mappings can be acquired from manually annotated parallel corpora (Padó & Erk, 2005). An interesting avenue for future work is to identify semantic role mappings in a fully automatic fashion.

## Acknowledgments

We are grateful to the three anonymous referees whose feedback helped to improve the present article. Special thanks are due to Chris Callison-Burch for the `linearb` word alignment user interface, to Ana-Maria Giuglea and Alessandro Moschitti for providing us with their shallow semantic parser, and to our annotators Beata Kouchnir and Paloma Krei-scher. We acknowledge the financial support of DFG (Padó; grant Pi-154/9-2) and EPSRC (Lapata; grant GR/T04540/01).

---

18. Note, however, that the cross-lingual projection of PropBank roles raises the question of their interpretation; see the discussion by Fung et al. (2007).